\documentclass[sigconf]{acmart}
\usepackage{graphicx} 
\usepackage{booktabs}
\usepackage{multirow}
\usepackage{subfigure}
\usepackage{amsmath}
\usepackage{balance}

\AtBeginDocument{%
  \providecommand\BibTeX{{%
    \normalfont B\kern-0.5em{\scshape i\kern-0.25em b}\kern-0.8em\TeX}}}

\copyrightyear{2022}
\acmYear{2022}
\setcopyright{acmlicensed}\acmConference[WSDM '22]{Proceedings of the
Fifteenth ACM International Conference on Web Search and Data
Mining}{February 21--25, 2022}{Tempe, AZ, USA}
\acmBooktitle{Proceedings of the Fifteenth ACM International Conference on
Web Search and Data Mining (WSDM '22), February 21--25, 2022, Tempe, AZ,
USA}
\acmPrice{15.00}
\acmDOI{10.1145/3488560.3498479}
\acmISBN{978-1-4503-9132-0/22/02}

\newtheorem{myDef}{Definition}

\DeclareUnicodeCharacter{2212}{-}
\begin{document}
\fancyhead{}

\title{Leaving No One Behind: A Multi-Scenario Multi-Task Meta Learning Approach for Advertiser Modeling}

\author{
  Qianqian Zhang, Xinru Liao, Quan Liu$^{*}$, Jian Xu, Bo Zheng} \thanks{$^*$Corresponding author: Quan Liu, lq204691@alibaba-inc.com}
\affiliation{Alibaba Group \country{China}}
\email{{qianfan.zqq, xinru.lxr, lq204691, xiyu.xj, bozheng}@alibaba-inc.com}

\begin{abstract}

Advertisers play an essential role in many e-commerce platforms like Taobao and Amazon. Fulfilling their marketing needs and supporting their business growth is critical to the long-term prosperity of platform economies. However, compared with extensive studies on user modeling such as click-through rate predictions, much less attention has been drawn to advertisers, especially in terms of understanding their diverse demands and performance. Different from user modeling, advertiser modeling generally involves many kinds of tasks (e.g. predictions of advertisers' expenditure, active-rate, or total impressions of promoted products). In addition, major e-commerce platforms often provide multiple marketing scenarios (e.g. Sponsored Search, Display Ads, Live Streaming Ads) while advertisers' behavior tend to be dispersed among many of them. This raises the necessity of multi-task and multi-scenario consideration in comprehensive advertiser modeling, which faces the following challenges: First, one model per scenario or per task simply doesn't scale; Second, it is particularly hard to model new or minor scenarios with limited data samples; Third, inter-scenario correlations are complicated, and may vary given different tasks. 

To tackle these challenges, we propose a multi-scenario multi-task meta learning approach (M2M) which simultaneously predicts multiple tasks in multiple advertising scenarios. Specifically, we introduce a novel meta unit that incorporates rich scenario knowledge to learn explicit inter-scenario correlations and can easily scale to new scenarios. Furthermore, we present a meta attention module to capture diverse inter-scenario correlations given different tasks, and a meta tower module to enhance the capability of capturing the representation of scenario-specific features. Compelling results from both offline evaluation and online A/B tests demonstrate the superiority of M2M over state-of-the-art methods.

\graphicspath{{fig/}}
\begin{figure}
   \centering 
   \vspace{-1pt}
   \includegraphics[scale=0.072]{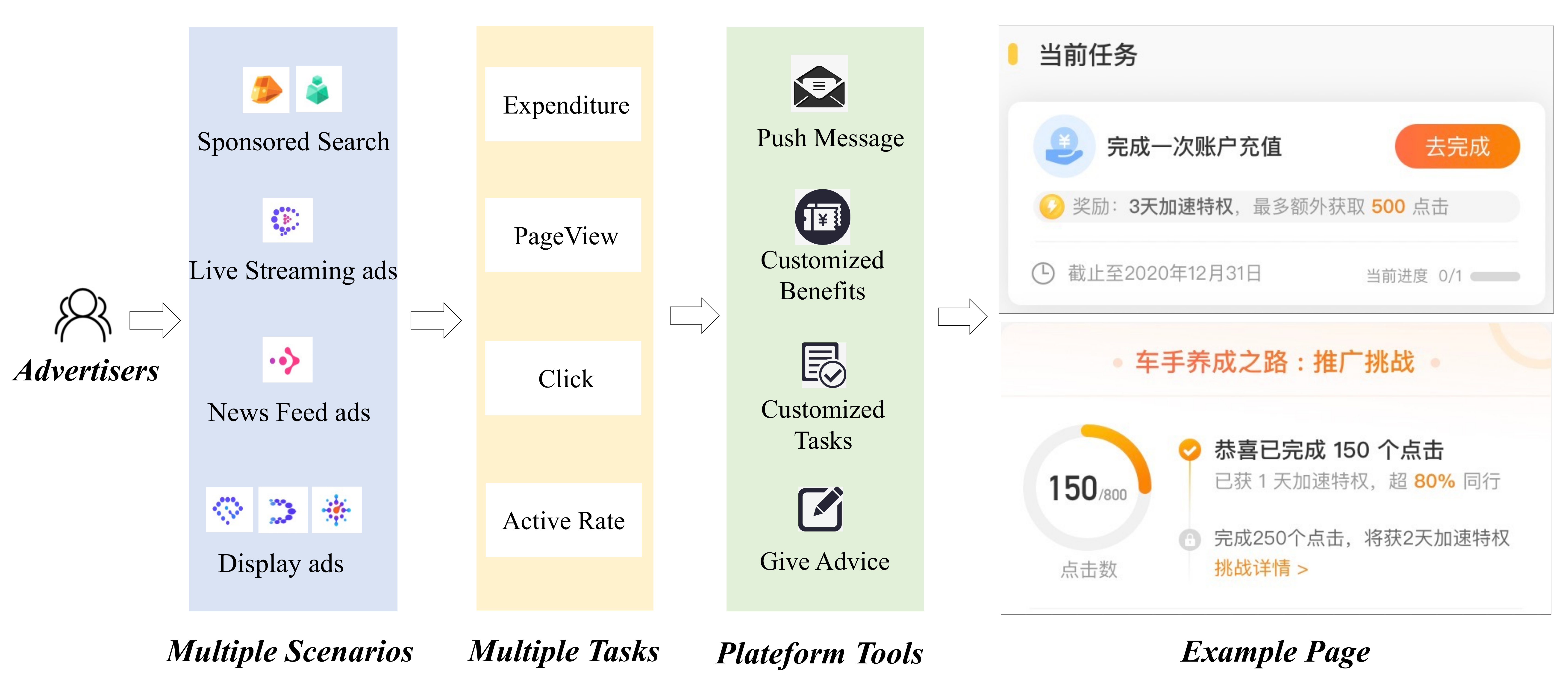}  

   \vspace{-0.1cm}
  \label{fig:advertiser modeling}
   \caption{ 
    The interaction flow between advertisers and the ad platform: 1) Advertisers operate in multiple scenarios; 2) Our model routinely generates multi-task predictions of each advertiser in corresponding scenarios; 3) These predictions feed into various advertising engagement tools.
    }
   \vspace{-0.1cm}
 \end{figure}

\end{abstract}



\maketitle
\section{Introduction}

In the Internet era, large e-commerce platforms such as Taobao and Amazon are becoming primary places for users to find and purchase products. With many potential consumers on e-commerce platforms, an increasing proportion of sellers are packed into e-commerce platforms to advertise their products. Although advertisers play an important role in this ecosystem, much less attention has been paid to understanding advertisers either from academic or industrial communities. Existing studies mainly focus on the user side
\cite{cheng2016wide, guo2017deepfm, zhou2019deep, zhou2018deep, li2019multi, pi2019practice, pi2020search},
while some \cite{guo2020deep}\cite{yoon2010prediction} noticed the necessity of advertiser understanding for platforms' long-term development, but they focused on predicting one single task like churn rate. As different advertisers at different business cycles have various demands as well as advertising performance, for example, impressions or clicks of advertised products, ROI (i.e return on investment), expenditure constraints, active or churn rate, etc, it is insufficient to measure overall conditions of advertisers based on one single task. 
Apart from multi-task demands, multi-scenario also needs to be considered in advertiser modeling. As illustrated in Figure 1, an advertiser may operate in multiple advertising scenarios such as sponsored search, news feed ads and even new emerging ones like live streaming ads (e.g. Taobao Live). A thorough modeling of advertisers require predicting advertisers' various tasks mentioned above in each scenario, without which we may end up in partial conclusions about advertisers. And this comprehensive understanding further enables advertising platforms to generate customized treatments like Rewarded Task and Push Message to better engage advertisers, which is quite essential for long-term development of platform economies.

\begin{table}[]
\caption{The pearson correlations of different pair of scenarios given different tasks.}
\label{tab:correlation}
\centering
\small
\setlength{\tabcolsep}{0.8mm}{
\begin{tabular}
{c|c|c}
\toprule
\textbf{Pearson Correlations}         & \textbf{Task.Expenditure}   & \textbf{Task.Click}           \\ \midrule
<Sponsored Search, News Feed Ads>           &   0.67  & 0.61                                 \\ \midrule
<Sponsored Search, Display Ads>           & 0.57  & 0.53 
\\ \midrule
<News Feed Ads, Display Ads>           & 0.70  & 0.45
\\ \bottomrule
\end{tabular}}
\end{table}

Modelling advertisers in a multi-scenario multi task fashion turns out to be far from trivial. In particular, it faces challenges including (1) \textbf{Limited reference samples in new or minor scenarios}. The number of training samples can be quite different for different scenarios, with some minor scenarios only having a few thousand daily active advertisers. A conventional solution would train exclusive models for each scenario, while the data from small scenarios are too sparse to train a reliable model. This problem worsens when predicting in new scenarios. How to transfer information among different scenarios while maintaining each's own specific characteristics is not trivial. (2) \textbf{Complex inter-scenario relations with multiple tasks}. Inter-scenario correlations are complex and not easy to precisely capture. Given a task like prediction of clicks or expenditure of advertisers in different scenarios, the inter-scenario correlations can sometimes be positive, negative, or unrelated. Furthermore, these correlations may evolve and vary when modeling different tasks. As illustrated in Table \ref{tab:correlation}, for the same task in different pair of marketing scenarios, the correlation coefficients are quite different. One common solution here is multiple task learning (MTL), which models the correlations of multiple tasks in one scenario, and learns simultaneously in one model and is proven to improve learning efficiency through information sharing between tasks \cite{ma2018modeling}\cite{tang2020progressive}\cite{misra2016cross}\cite{liu2020multi}\cite{liu2020incorporating}. However, these MTL-based methods cannot explicitly or effectively model multiple scenarios beyond multiple tasks. How to capture the complex inter-scenario correlations with multiple tasks to improve prediction performance of multiple scenarios is another key challenge. 

To address these challenges, we propose M2M, a multi-scenario multi-task meta learning model, to predict multiple tasks in multiple scenarios all in a fully automatic manner. First, a backbone network is adopted to learn representations of advertiser-related features and task representations. Then, to learn the meta-knowledge of scenario-related information explicitly, we design a Meta Unit with customized scenario-related input, and its output is used to generate layer weights for later prediction networks. Furthermore, to combine multiple-scenario multiple-task in an end-to-end fashion effectively, we propose a Meta Learning Mechanism, which comprises a meta attention layer and a meta residual tower layer, respectively to capture diverse inter-scenario correlations and enhance scenario-specific feature representation capability. Finally, we utilize the joint Poisson loss for multiple tasks to optimize our model in an end-to-end manner. Compelling results from both offline evaluation and online A/B tests demonstrate the superiority of M2M over state-of-the-art methods, especially for long-tail scenarios.

The contributions of this paper are highlighted as follows:
\begin{itemize}
    \item  We study the important but under-explored problem of comprehensive advertisers modeling to engage advertisers and help them grow, which is essential to the prosperity of advertising platforms. To the best of our knowledge, we are the first attempt to explicitly use different scenario-related information and propose a unified framework for advertiser understanding.
    \item 
    We propose M2M, a novel multi-scenario multi-task paradigm, to predict various tasks from various scenarios in an end-to-end fashion. The key idea is that we combine the advantage of meta learning and multi-task learning,  which enhances the ability of capturing diverse inter-scenario correlations given different tasks. Moreover, this approach easily scales to new scenarios and is proven to significantly improve the performance in new or minor scenarios.
    \item Compelling results from both offline evaluation and online A/B tests demonstrate the superiority of M2M over state-of-the-art methods, especially for emerging and minor scenarios. And our model has been successfully deployed in various advertising-related applications at Taobao.
\end{itemize}

\section{Related work}

\graphicspath{{fig/}}
 \begin{figure}
   \centering 
   \vspace{-4pt}
   \includegraphics[scale=0.22]{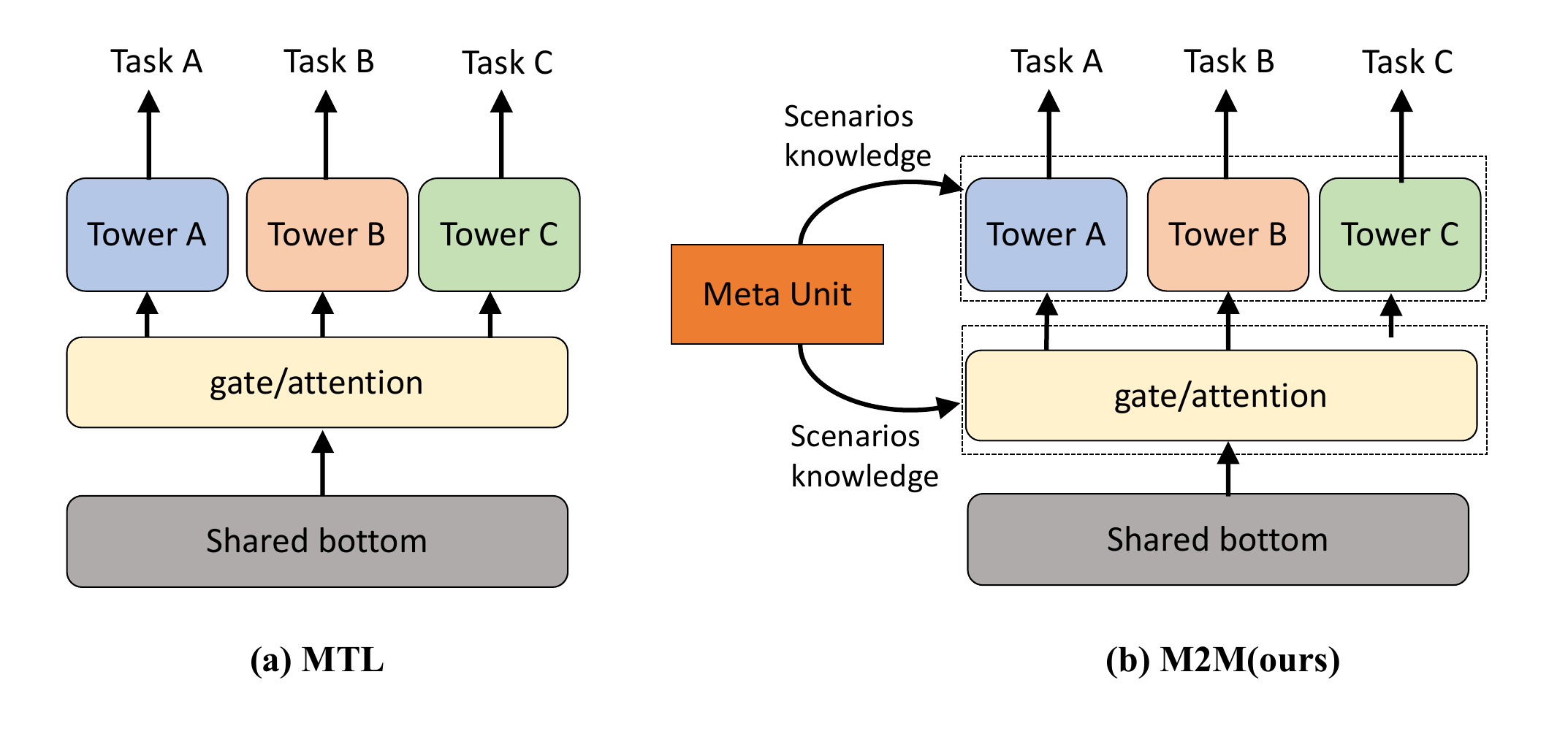}  
   \vspace{-10pt}
   \caption{ 
    Different Solution: Figure(a) is the conventional deep model of multi-task learning which incorporates shared-bottom expert structures and gating network to learn expert representations, and feeds into a specific tower network. Figure(b) additionally uses meta unit of scenario knowledge to generate network weights for different scenarios. }\label{fig:difsolution}
   \vspace{-1pt}
 \end{figure}

\textbf{User Modeling.} In recommendation systems and digital advertising space, there is extensive work in modeling user intents and satisfaction like CTR(click-through rate)\cite{zhou2018deep}\cite{pi2019practice}\cite{han2021joint}\cite{han2019graphconvlstm} and CVR(post-click conversion rate), which is crucial for matching and ranking systems. However, there is much less work on modeling advertisers' various aspects like impressions, clicks, GMV, active-rate, etc. \cite{yoon2010prediction} presents a method that carefully selects a homogeneous group of advertisers and uses classification algorithms to predict advertisers' churn. \cite{guo2020deep} proposes a novel two-stage network to model advertiser intent and satisfaction simultaneously to provide better services for advertisers, but scenarios related information was not considered. Compared with user intent, modeling advertisers' behavior and performance is actually much more complicated for several reasons. First, user intents are generally more explicit, represented by clicks and purchases, while advertiser intents are harder to identify due to larger variety. For example, some may aim at controlling costs while others thrive for more impressions and sales. Second, user-side usually has rich data and uniformed behavior, while advertisers' operations are more sparse and diverse.

\textbf{Multi-task Learning.} Multi-task learning\cite{2017An}\cite{zhang2017survey} is a machine learning paradigm that aims to leverage useful information across multiple tasks to improve performance of each task, which has many applications in various areas such as natural language processing\cite{sogaard2016deep}, computer vision\cite{kendall2018multi}\cite{Collobert2008A} and speech recognition\cite{deng2013new}\cite{2016Multi}. In early literature of multi-task learning, \cite{evgeniou2005learning} proposes a kernel method and regularization model to learn multi-task simultaneously. In the deep learning era, various deep representation learning approaches have been proposed for multi-task learning problems. \cite{thrun2012learning} proposes to share parameters of hidden layers which is a commonly used approach. And \cite{misra2016cross} proposes a cross-stitch unit that learns to combine task-specific layer for each task linearly. \cite{ma2018modeling} proposes Multi-gate Mixture-of-Experts(MMoE) to model the relationship across different tasks by sharing experts at the bottom layer, while utilizing different gate networks for each task to obtain fusing information of experts. CGC\cite{tang2020progressive} develops further on the basis of MMoE, separating task-common and task-specific parameters explicitly to avoid parameter conflict resulting from complex task correlations. \cite{kendall2018multi} defines variance uncertainty in loss to optimize the weights in multi-task learning.
Despite their respective success in capturing
multiple tasks relationship, to the best of our knowledge, few of these works considers modeling scenarios information explicitly.



\textbf{Meta Learning.} Meta-learning\cite{hospedales2020meta}\cite{peng2020comprehensive}\cite{schmidhuber2007godel}\cite{thrun2012learning}\cite{vanschoren2018meta}, or learning to learn, is the science of systematically observing how different machine learning approaches perform in multiple tasks, and learning from this experience to seek rapid and accurate model adaptation to unseen tasks, even with few training examples. It has been widely used in many applications like highly automated AI, few-shot learning, natural language processing\cite{sogaard2016deep} and image processing\cite{finn2017model}\cite{munkhdalai2017meta}. Meta-learning can help improve transfer learning and multi-task learning by using a meta-objective. MAML\cite{finn2017model} and Reptile\cite{2018On} propose an optimization-based and model-agnostic algorithm that uses meta-gradient updates to train the parameters explicitly so that a small number of gradient steps produces good generalization performance on new tasks. \cite{munkhdalai2017meta} introduces a novel MetaNet structure, which consists of two main learning components: the base learner performs in the input task space whereas the meta learner operates in a task-agnostic metaspace. \cite{ha2016hypernetworks}explores a more efficient method that uses a smaller network named hyper network to generate the structure of weights for a more extensive network, in order to search within the much smaller weight space.\cite{bertinetto2016learning} constructs an efficient feed-forward one-shot learner, which demonstrates that a deep neural network can learn at the “meta-level” of predicting filter parameters for another main network. ST-MetaNet \cite{pan2019urban} employs sequence-to-sequence architecture, leveraging the meta-knowledge extracted from geography attributes to generate parameter weights for a graph attention network that captures diverse spatial correlation and a recurrent network that considers diverse temporal correlation. Unlike above works, we explicitly consider multi-level scenario information in one model, and combine the advantage of multi-task learning and meta learning as shown in Figure \ref{fig:difsolution}. 


\section{Preliminaries}



\subsection{Problem Settings}
\begin{table}[]
\caption{Feature types and descriptions}
\label{tab:feature}
\small
\setlength{\tabcolsep}{1mm}{
\begin{tabular}{@{}c|c@{}}
\toprule
\textbf{Feature Type}         & \textbf{Description}                                                                                          \\ \midrule
Scenario Attributes            & Scenario Type, Statistic Scenario Characteristics, etc.                                 \\ \midrule
Advertiser Profile      & Business Scale,  Main Category, Star Level, etc.
\\ \midrule
Multi-type Behavior      & Login, Campaign Modification, Bidding, Budget, etc.
\\ \midrule
Multi-type Performance  & Expenditure, GMV, ROI, impressions etc.
\\ \bottomrule
\end{tabular}}
\end{table}

Advertisers accumulate multiple types of behavior and performance in different advertising scenarios. To fully represent advertisers, we also establish advertiser profile and summarize scenario knowledge. More details about our used features are shown in Table \ref{tab:feature}.


\begin{myDef}
\textbf{Scenario Attributes}. Let $\mathbf{S}=\{{s}_1, {s}_2, \dots, {s}_{l}\}$ denote the set of scenarios related information which is advertisers' marketing scenario like sponsored search, display ads, etc.
\end{myDef}

\begin{myDef}
\textbf{Advertiser Profile}. Let $\mathbf{A}=\{{a}_1, {a}_2, \dots, {a}_{m}\}$ denote the set of advertiser profile information.
\end{myDef}

\begin{myDef}
\textbf{Multi-type Behavior Sequences}. Let $\mathbf{X}_{b}=\{\mathbf{X}^{t}_{b}\}^T_{t=1}$ denote time-dependent multi-type behavior information in a time period $T$, where $\mathbf{X}^{t}_{b}=\{\mathbf{x}^{t}_{b1}, \mathbf{x}^{t}_{b2}, \dots, \mathbf{x}^{t}_{bm}\}$ denote dense or sparse features of advertisers at time step $t$.
\end{myDef}

\begin{myDef}
\textbf{Multi-type Performance Sequences}. Let $\mathbf{X}_{p}=\{\mathbf{X}^{t}_{p}\}^T_{t=1}$ denote time-dependent multi-type performance information in a time period $T$, where  $\mathbf{X}^{t}_{p}=\{\mathbf{x}^{t}_{p1}, \mathbf{x}^{t}_{p2}, \dots, \mathbf{x}^{t}_{pm}\}$ denote dense or sparse features of advertisers at time step $t$.
\end{myDef}

\textbf{Multi-Scenario Multi-Task Prediction}. 
Given scenario attri-butes $\mathbf{S}$, advertiser profile $\mathbf{A}$, multi-type behavior sequences $\mathbf{X}_{b}$, multi-type performance sequences $\mathbf{X}_{p}$, we need to train a unified model, to predict multiple tasks (e.g. expenditure, active-rate, clicks) in multiple scenarios(e.g. Sponsored Search, Display Ads, Star Shop) for the next $\tau$ days, which can be formulated as:
\begin{equation}
\hat{\mathbf{Y}}^{t+\tau}_{ij}\mid _{scenarios_{i},task_{j}} = \mathbf{\Gamma}(\mathbf{S},\mathbf{A},\mathbf{X}_{b},\mathbf{X}_{p})
\end{equation}
where d is the function we aim to learn, note that the proposed method can be easily extended to any time interval prediction.

\section{THE PROPOSED M2M MODEL}

\graphicspath{{fig/}}
 \begin{figure}
  \centering 
  \vspace{1pt}
  \includegraphics[scale=0.115]{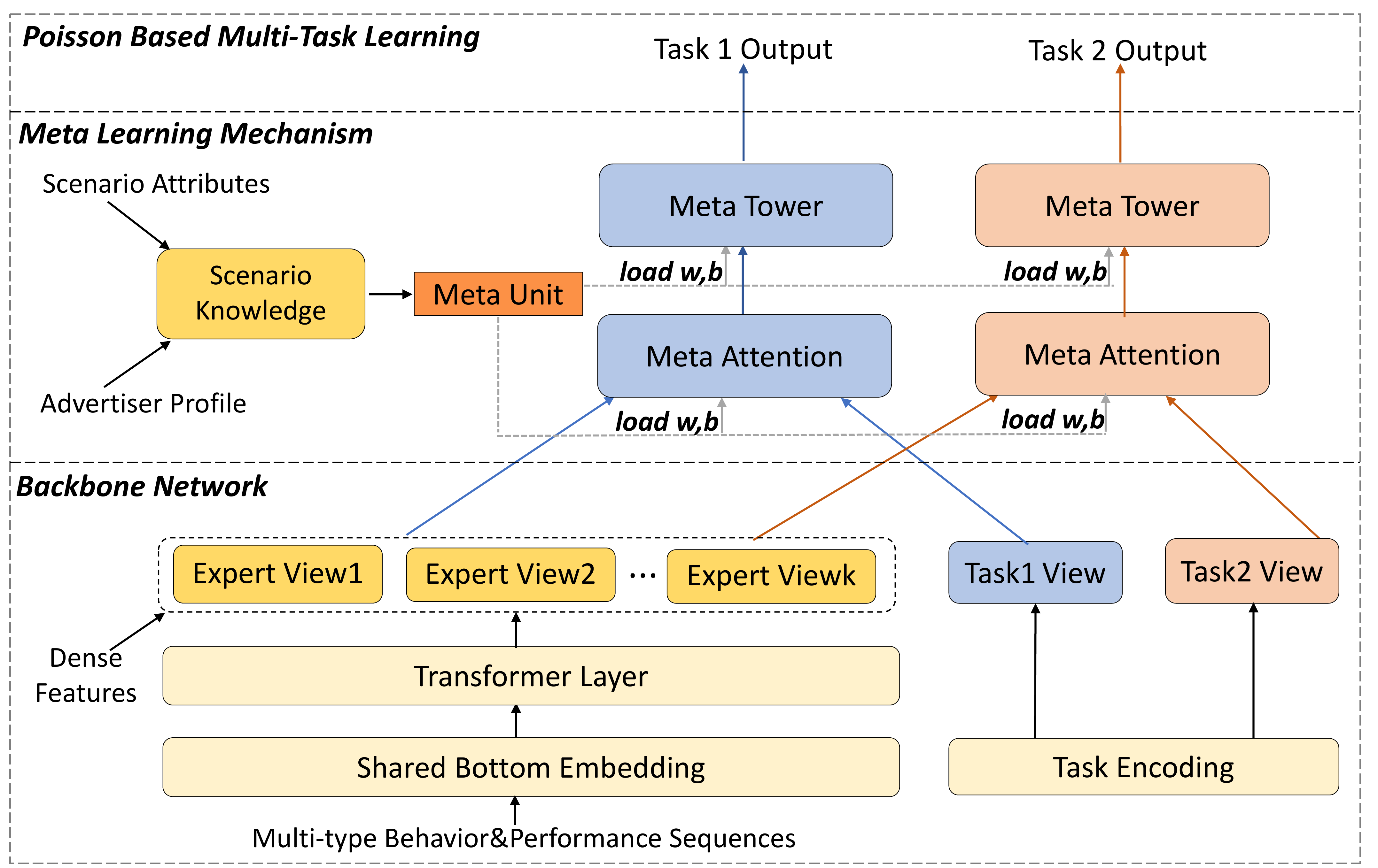}  
  \vspace{0pt}
  \caption{ 
    Overview of M2M Framework. The backbone network aims to obtain feature and task representations. The meta learning mechanism uses rich scenario knowledge to generate meta network weights of meta attention module and meta tower module. And the poisson-based multi-task learning uses poisson regression to generate predictions.}\label{fig:overview}
  \vspace{0pt}
 \end{figure}

In this section, we introduce our Multi-Scenario Multi-Task advertiser Modelling(M2M) approach which consists of three main stages. The first stage, \textit{Backbone Network}, aims to obtain feature and task representations. The second stage \textit{Meta Learning Mechanism} comprises a meta attention module and meta tower module, to capture diverse inter-scenario correlations and enhance scenario-specific feature representation capability respectively. The third stage \textit{Poisson Based Multi-Task Learning} is introduced to predict multiple tasks from multiple scenarios by Poisson loss. The overall architecture is depicted in Figure \ref{fig:overview}. 

\subsection{Backbone Network}

\subsubsection{Shared Bottom Embedding}
Since these input sequence features $\{\mathbf{X}_{b},\mathbf{X}_{p}\}$ include continuous features like the page view, click, or expenditure of different marketing scenarios, the first step is discretization. After discretization, the inputs are high dimensional binary vectors. We use embedding layer to transform them into low-dimensional dense representations. Finally, we obtain the embedding with fixed-size low-dimensional vectors of each time sequence, and then we can use the transformer layer to embed these performance and behavior sequence features 
separately. Note that we also use positional embedding for each time sequence to capture the order or time information in historical 
performance or behavior sequence. Here we concatenate the trainable position embedding into each time sequence embedding feature.
\subsubsection{Transformer Layer}
After embedding these features into low dimensional representations, we use transformer layer to learn the deeper representation of each time sequence by capturing the relations with other time sequence features, which are more computationally efficient and expressively more powerful compared with RNN and LSTM\cite{hochreiter1997long}. \\
\textbf{Self-attention Layer}. We use the scaled dot-product attention which is defined in \cite{2017Attention}.
\begin{equation}
Attention(\mathbf{Q},\mathbf{K},\mathbf{V}) = softmax(\frac{\mathbf{Q}\mathbf{K^\intercal}}{\sqrt{d}}\mathbf{V}),
\end{equation}
where $\mathbf{Q}$ represents the query, $\mathbf{K}$ the key, $\mathbf{V}$ the value of input embedding, and $\mathbf{d}$ the scale factor is the dimension of the key. The three matrices are the equal-dimensional linear projects of the input time sequence embedding features. Following \cite{2017Attention} we use multi-head self-attention:
\begin{equation}
MH(\mathbf{X}) = concat(head_{1},head_{2},...,head_{h})\mathbf{W^H},
\end{equation}
\begin{equation}
head_{i} = Attention(\mathbf{X}\mathbf{W_i^Q},\mathbf{X}\mathbf{W_i^K}, \mathbf{X}\mathbf{W_i^V}),
\end{equation}
where h is the number of heads, $\mathbf{X}$ the processed input embedding features of performance or behavior, $\mathbf{W_i^Q}$, $\mathbf{W_i^K}$ and  $\mathbf{W_i^V}$ are the project matrices and trainable parameters of $head_i$ respectively. Note that since we have the following expert view layer explained in 4.1.3, we doesn't incorporate additional none-linear units in the transformer layer. After we use transformer layer to learn to representation of behavior and performance sequence feature respectively, we concatenate then into a fixed-length vector.
\begin{equation}
\mathbf{F} = concat(MH(\mathbf{X_b}),MH(\mathbf{X_p}))
\end{equation}

\subsubsection{Expert View Representation}
After the transformer layer, we obtain the representation of the input feature $\mathbf{F}$, which we can also concatenate other dense feature of advertisers. And following MMoE network structure, we build mixture-of-experts layers which obtain different shared representations.

\begin{equation}
\mathbf{E_{i}}=f_{MLP}(\mathbf{F}), \forall{i} \in 1,2,..,k
\end{equation}
where $\mathbf{E_{i}}$ the $i$th expert output, k the number of expert views.

\subsubsection{Task View Representation}

Besides embedding features, inspired by MRAN\cite{zhao2019multiple}, we also embed all the tasks in the same space, which act as the “anchor points” of the tasks and introduce prior knowledge of tasks and influence the weight of feature information. Note that there is no specific label information in the test set. Therefore, the task representation we extract is a global one rather than local. We first use the lookup table to transform one-hot task information into low dimensional dense representations; Then, a feed-forward layer is adopted to reshape the dimension. The feed-forward sub-layer is composed of a nonlinear layer with LeakyReLU\cite{leakyrelu} activation. Formally, we have the following equation :
\begin{equation}
\mathbf{T}_{t}  = f_{MLP}(Embedding(t)), \forall{t} \in 1,2,..,m
\end{equation}
where $\mathbf{T}_{t}$ denotes the anchor embedding of task t, m the number of tasks.

\subsubsection{Scenario Knowledge Representation}
Scenario knowledge representation is the knowledge representation of the advertiser's advertising scenarios. Here we not only use the scenario attributes that the advertiser marketing, but also the advertiser profile information to better learn the multiple tasks. We incorporate a feed-forward layer to learn to representation of scenario knowledge.
\begin{equation}
\mathbf{\Tilde{S}}  = f_{MLP}(\mathbf{S},\mathbf{A})
\end{equation}

\subsection{Meta Learning Mechanism}

\begin{figure*}
   \centering 
   \vspace{-4pt}
   \includegraphics[scale=0.18]{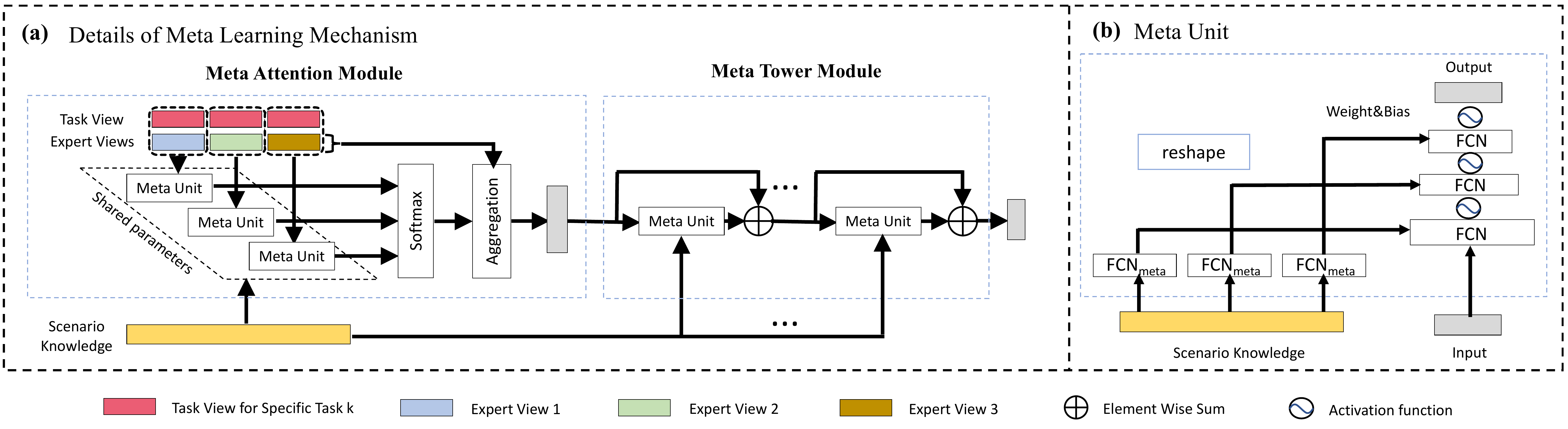}  
   \vspace{-8pt}
   \caption{ 
   The Meta Learning Mechanism of our proposed M2M Framework. The figure shows the pipeline of one task which contains meta attention module and meta tower module. The meta attention module combines corresponding task view and expert views using network weights generated by specific scenario knowledge to get weighted representations of expert views. The meta tower module here uses residual tower network whose weights are also generated by corresponding scenarios knowledge.}\label{fig:detail}
   \vspace{-2mm}
 \end{figure*}

To better characterize the scenario-specific representation from different sequential features, we propose a meta learning mechanism that consists of two components: a meta attention module and a meta residual tower module. These two modules are hierarchically organized. The meta attention module is positioned lower to capture diverse inter-scenario correlations, and a meta residual tower module is placed upper to enhance the capability in capturing the scenario-specific feature representation, as shown in Figure \ref{fig:detail}.

\graphicspath{{fig/}}

\subsubsection{Meta Unit}

In our meta learning mechanism, we propose a meta unit to model different scenario-related information explicitly. As shown in Figure \ref{fig:detail}, to better capture inter-scenario correlations, we use scenarios knowledge $\mathbf{\Tilde{S}}$ as the input of meta unit. The meta unit transforms the scenario knowledge into the dynamic weights and bias parameters for meta attention learning and meta tower learning, which can be formulated as :
\begin{equation}
\mathbf{h}^{(0)} = \mathbf{h}_{input}, 
\end{equation}
\begin{equation}
\mathbf{h}^{(i)} = \sigma(\mathbf{W}^{(i-1)} \mathbf{h}^{i-1}+\mathbf{b}^{(i-1)}),  \forall{i} \in 1,2,..,K
\end{equation}
where $\mathbf{h}_{input}$ is the input vector with dimension $d$, $\sigma$ is a non-linear activation function. K is the total layer number of meta unit. $\mathbf{W}^{(i)}\in\mathcal{R}^{\mathbf{d}\times\mathbf{d}}$ and $\mathbf{b}^{(i)}\in \mathcal{R}^\mathbf{d}$ are projection parameters. In particular, these projection parameters are generated by the meta unit from the scenario knowledge as shown in Figure \ref{fig:detail}. Formally:
\begin{equation}
\mathbf{W}^{(i-1)} = Reshape(\mathbf{V}_{w} \mathbf{\Tilde{S}}+\mathbf{v}_{w}),
\end{equation}
\begin{equation}
\mathbf{b}^{(i-1)} = Reshape(\mathbf{V}_{b} \mathbf{\Tilde{S}}+\mathbf{v}_{b}),
\end{equation}
Where $\mathbf{\Tilde{S}}$ represents scenario knowledge, after a fully connected network, a $Reshape$ function is adopted to obtain weight matrix $\mathbf{W}^{(i)}$ and bias vector $\mathbf{b}^{(i)}$. 
\begin{equation}
\mathbf{h}_{output} = \mathbf{h}^{K} = Meta(\mathbf{h}_{input}), 
\end{equation}
Finally, the output of meta unit is obtained as $\mathbf{h}_{output}$, we formally define the meta unit process as a $Meta$ function.

Different from conventional works that add scenario related information as input features to models, our meta unit explicitly models inter-scenario correlations by incorporating scenario knowledge and producing dynamic parameter weights for each specific scenario. Through obtaining meta-representation of specific scenarios, the meta unit is proven to significantly improve the expressive power in new or minor scenarios.

\subsubsection{Meta Attention Module}

The meta attention module at the bottom tries to learn the contributing weights of feature views derived from the backbone network. Intuitively, each task may rely on different parts of the shared feature views. However, it is inappropriate to directly calculate the attention score ignoring the scenario factors. For instance, the amount of training data may differ for different scenarios, and the data distribution of different scenarios may be dynamic for a specific task. While traditional alignment attention module can model the correlation of tasks and feature views, scenarios-specific diversity tends to be overlooked. Hence, we propose augmenting our M2M framework to capture the scenario signals while calculating the attention score, which can help the attention module learn how to produce dynamic attention weight in different scenarios. In specific, to model different patterns of attention score given the scenarios related information, different from traditional concatenation-based attention, we employ the above section meta unit, which is formulated as :
\begin{equation}
a_{t_i} = \mathbf{v}^{T}Meta_t([\mathbf{E_{i}}\parallel\mathbf{T}_{t}]),
\end{equation}
\begin{equation}
\alpha_{t_i} =\frac{exp(a_{t_i})}{\sum^{M}_{j=1}exp(a_{t_j})},
\end{equation}
where $\mathbf{E_{i}}$ is the sub-view feature vector of dimension $d_1$, $\mathbf{T}_{t}$ is the embedding of task anchor embedding of dimension $d_2$, $\sigma$ is a non-linear activation function. The hidden vector $\mathbf{v}\in\mathcal{R}^{(d_1+d_2)}$ projects the hidden vector into a scalar weight for output.

The final representation $R_{t}$ for corresponding task ${t}$ is the summation of the multiple sub-views weighted by their attention weights:
\begin{equation}
\mathbf{R}_{t} = \sum^{k}_{i=1}\alpha_{t_i}{\mathbf{E}_{i}}
\end{equation}
Since we extract scenario knowledge and use such information to generate the weights and bias of the meta attention module, this module not only can model tasks and features' relevance, but also has the ability to capture the complex inter-scenario correlations.

\subsubsection{Meta Tower Module}
After obtaining the aggregated latent representations from the meta attention module, our M2M adopts a meta residual tower module\cite{2016Deep} which also contains meta unit to distinguish different scenarios. Since feature patterns may vary from scenario to scenario for a specific task, a simple shared feed-forward network is not sufficient to capture diverse scenarios information. To model such diversity, we introduce meta residual tower module as follows :
\begin{equation}
\mathbf{L}^{(0)}_t = \mathbf{R}_{t},
\end{equation}
\begin{equation}
\mathbf{L}^{(j)}_t = \sigma(Meta^{(j-1)}(\mathbf{L}^{(j-1)}_t)+\mathbf{L}^{(j-1)}_t), \forall{j} \in 1,2,..,L
\end{equation}
where $\mathbf{R}_{t}$ is the attention output of sub-views from different views, $\sigma$ is a non-linear activation function. $L$ is the number of residual layer.

\begin{table}
\caption{Dataset Details.}
\label{tab:dataset}
\small
\setlength{\tabcolsep}{0.5mm}{
\begin{tabular}{@{}c|ccccc@{}}
\toprule
\textbf{Dataset}                          & \textbf{Sce.A}                 & \textbf{Sce.B}                & \textbf{Sce.C}              & \textbf{Sce.D}                & \textbf{Sce.E}                   \\ \midrule
\multicolumn{1}{l|}{\textbf{Description}} & \multicolumn{1}{l}{Sponsor Search} & \multicolumn{1}{l}{Display} & \multicolumn{1}{l}{Star Shop} & \multicolumn{1}{l}{News Feed} & \multicolumn{1}{l}{Brand Search} \\ \midrule
\textbf{Train Data}                       & 25.28M    & 0.51M   & 0.47M          & 7.29M   & 0.84M                                     \\ \midrule
\textbf{Valid Data}                       & 5.36M     & 0.12M       & 0.11M          & 1.50M   & 0.16M                                \\ \midrule
\multicolumn{1}{l|}{\textbf{Test Data}}   & 5.21M       & 0.11M    & 0.10M          & 1.52M   & 0.15M                                     \\ \bottomrule
\end{tabular}}
\end{table}

\subsection{Algorithm of Optimization}

To train M2M jointly in an end-to-end manner, we adopt a common formulation of joint loss by calculating the weighted sum of the losses for each individual task with input $\mathbf{X}$ and task-specific labels $\mathbf{Y}^{t}\in\mathbf{Y},t = 1,2,...,m$,which is defined as :
\begin{equation}
\begin{split}
\mathcal{L}_{train}(\mathbf{S},\mathbf{A},\mathbf{X_b},\mathbf{X_p},\mathbf{Y})=&\sum^m_{t=1}\lambda_t \mathcal{L}^{t}(\mathbf{S},\mathbf{A},\mathbf{X_b},\mathbf{X_p},\mathbf{Y}^t)+
\\
&\alpha (\left \| \mathbf{W}_1 \right \|_{F}^2 + \left \| \mathbf{W}_2 \right \|_{F}^2)\\
\end{split}\end{equation}

where the first term is a loss function, which is a linear combination of task-specific losses $\mathcal{L}^{t}$ with task weightings $\lambda_t$. The latter term is to penalize the complexity of $\mathbf{W}_1$ and $\mathbf{W}_2$ which denotes parameters from meta unit and other network weights respectively, and $\alpha$ is the regularization weight. 

Poisson loss [8] is a widely used loss function for discrete count data, for each task $t$, we optimize the following objective function:
\begin{equation}
\mathcal{L}^{t}(\mathbf{S},\mathbf{A},\mathbf{X_b},\mathbf{X_p},\mathbf{Y}^t) = \frac{1}{N}\sum^N_{i=1}(\hat{y}^{t}_{i}-y^{t}_{i}log\hat{y}^{t}_{i})
\end{equation}






\begin{table*}[]
\caption{Comparison with State-of-the-arts. Sce.a represents Sponsor Search scenario; Sce.b represents Display scenario which is a \underline{minor scenario}; Sce.c represents Star Shop scenario which is a \underline{new scenario}. M2M outperforms all due to 2 key advantages: the well designed Meta Unit, and the effective meta learning mechanism that captures diverse inter-scenario correlations given different tasks. 
}
\label{tab:Performance comparison}
\small
\setlength{\tabcolsep}{0.5mm}{
\begin{tabular}{@{}c|cc|cc|cc|cc|cc|cc|cc@{}}
\toprule
\multirow{2}{*}{Method} & \multicolumn{2}{c|}{Task.click(Sce.a)} & \multicolumn{2}{c|}{Task.click(Sce.b)} & \multicolumn{2}{c|}{Task.click(Sce.c)} & \multicolumn{2}{c|}{Task.active(Sce.a)} & \multicolumn{2}{c|}{Task.active(Sce.b)} & \multicolumn{2}{c|}{Task.active(Sce.c)} & \multicolumn{2}{c}{Overall(5Task*5Sce)} \\
                        & NMAE               & SMAPE             & NMAE               & SMAPE             & NMAE               & SMAPE             & NMAE               & SMAPE              & NMAE               & SMAPE              & NMAE               & SMAPE              & NMAE                & SMAPE              \\ \midrule
Single Task             & 0.2398             & 0.4684            & 0.6687             & 0.8354            & 0.8205             & 0.8971            & 0.2603             & 0.2345             & 0.4146             & 0.3501             & 0.3653             & 0.3620             & 0.5372              & 0.6349             \\
Shared Bottom           & 0.2740             & 0.4908            & 0.7486             & 0.8406            & 0.9388             & 0.9349            & 0.3112             & 0.2693             & 0.3897             & 0.3341             & 0.3246             & 0.3401             & 0.5561              & 0.6411             \\
Cross-Stitch            & 0.2474             & 0.4993            & 0.6657             & 0.8241            & 0.7621             & 0.8902            & 0.3612             & 0.3112             & 0.3703             & 0.2617             & 0.2820             & 0.2745             & 0.4984              & 0.6189             \\
MMoE                    & 0.2449             & 0.4711            & 0.6869             & 0.8085            & 0.7306             & 0.8429            & 0.3195             & 0.2674             & 0.3631             & 0.2686             & 0.2711             & 0.2649             & 0.4953              & 0.6101             \\
MRAN                    & 0.2254  & 0.4566 & 0.6668 & 0.7797 & 0.7103 & 0.8185 & 0.3252 & 0.2686 & 0.3735 & 0.2707 & 0.2703 & 0.2654 & 0.4921 & 0.6068 \\
CGC                     & 0.2277             & 0.4765            & 0.6235             & 0.8043            & 0.7377             & 0.8351            & 0.2714             & 0.2578             & 0.3394             & 0.2721             & 0.2476             & 0.2550             & 0.4879              & 0.6087             \\ \midrule
M2M                     & \textbf{0.2168}    & \textbf{0.4435}   & \textbf{0.4733}    & \textbf{0.7605}   & \textbf{0.5406}    & \textbf{0.7611}   & \textbf{0.2354}    & \textbf{0.2167}    & \textbf{0.3063}    & \textbf{0.2066}    & \textbf{0.2158}    & \textbf{0.2068}    & \textbf{0.4109}     & \textbf{0.5722}    \\ \bottomrule
\end{tabular}
}
\vspace{3ex}
\end{table*}

\section{Experiments}

In this section, we conduct extensive experiments to verify the effectiveness of our proposed M2M framework. Our experiments are designed to answer the following questions: \textit{1) What is the performance of our proposed M2M compared with other state-of-art methods? 2) What is the influence of each of M2M’s components? 3) What is the effect of key hyper-parameters, such as the length of temporal sequence, the dimension of scenario knowledge, the number of layers in meta unit? 4) What is the online A/B test performance of our proposed M2M? }


\subsection{Experimental Settings}

\subsubsection{Dataset Statistics.}
We conducted experiments on a large-scale industrial dataset collected from Taobao, ranging from 2019-12-10 to 2020-12-30. There are 5 marketing scenarios including sponsored search, display ads, star shop ads, news feed ads, and brand search ads denoted from Sce.A to Sce.E, with a total of 1.83 million advertisers, and 30 million samples in the data set as shown in Table \ref{tab:dataset}. As mentioned above, future PV(Page View or impressions), Click, Expenditure, GMV, Active rate are typical tasks that our model predicts in the dataset. In this evaluation, we used previous 40-day features to predict the next 7-day performance, and partitioned the data based on the time axis into non-overlapping training, evaluation, and test data, by a ratio of 8:1:1. It is noteworthy that the interval between training set and testing set should be no less than one week in case of data leakage. The dataset's statistical information is exhibited in table \ref{tab:dataset}.



\subsubsection{Comparative Methods}
We compare M2M with the following baselines:
\begin{itemize}
    \item \textbf{Single-task model}. Use the Embedding\&MLP architecture as the basis single-task deep learning model.
    \item \textbf{Shared Bottom}. The shared bottom model shares the bottom layer in multi-task learning.
    \item \textbf{Cross-Stitch Network}. Cross-Stitch\cite{misra2016cross} uses a linear cross-stitch unit to learn to optimally combine multiple network of representations for each task. 
    
    \item\textbf{MMoE}. MMoE\cite{ma2018modeling}
    makes use of a shared module named expert to learn the underlying feature representations. Each task uses its own gate network to combine experts to capture correlations and differences between the tasks.
    \item \textbf{MRAN}. MRAN\cite{zhao2019multiple} contains a task-feature dependence learning module, which can associate the related features with target tasks separately.
    \item \textbf{CGC}. CGC\cite{sogaard2016deep} consists of task-specific experts and the shared experts in representation layer, which avoids parameter conflicts resulting from complex task correlations.
\end{itemize}

\subsubsection{Implementation Details.}

For fair comparison, we search for optimal parameters on validation data and evaluate these models on test data; Besides, scenario related information is also added as the input features to base models. In the shared bottom embedding layer of M2M framework, we set the hidden state dimension $\mathbf{d_{input}}$ as 16. In the multi-behavior transformer module, we set the number of attention heads for multi-dimensional learning as 2. To better characterize the relationship of features and tasks, we introduce feature views and task views representation, all feature views dimensions and task views dimensions are set to $\mathbf{d_{view}}$ = 256. We use LeakyReLU\cite{leakyrelu} as the activation function for all models. During training, the batch size is set to 256. We use Adam optimizer \cite{adam} with the setting $\beta 1$ = 0.9, $\beta 2$ = 0.998 and $\varepsilon  = 1 \times 10^{- 9}$. The learning rate is set to $2 \times 10^{- 3}$. Gradient clipping is applied with range [−3, 3]. The model is trained in 10 epochs.

All the above models are trained using XDL\footnote{https://github.com/alibaba/x-deeplearning} platform. The trained parameters are distributed on 40 workers (15 CPU cores for each worker) and updated asynchronously.

\subsubsection{Evaluation Metrics.}
Commonly used metrics for the evaluation of regression problems are MAPE (Mean Absolute Percentage Error) which evaluates the micro average performance, and NMAE (Normalized Mean Absolute Error) which evaluates the macro average performance. In our tasks, the label of pv, clicks, etc, may be zero which would result in undefined value of MAPE. Therefore, we adopt SMAPE (Symmetric Mean Absolute Percentage Error), which is calculated as follows :
\begin{equation}
    SMAPE = \frac{1}{N}\sum_{i=1}^N\frac{\left | \hat{y_i}-y_i \right |} {(\hat{y_i}+y_i)/2}
\end{equation}
And NMAE metric is calculated as follows:
\begin{equation}
    NMAE = \frac{\sum_{i=1}^N\left | \hat{y_i}-y_i \right |}{\sum_{i=1}^N y_i}
\end{equation}
\\where N is the number of advertisers, $\hat{y_i}$ and ${y_i}$ are the predicted tasks and ground-truth of i-th advertiser, respectively.


\begin{figure}
    \begin{minipage}{1.0\linewidth}
        \centering   
        \vspace{-8pt}
        \subfigure[{\footnotesize Overall Performance in NMAE}]{\label{fig:jonitnpre_1}
        \includegraphics[width=0.48\textwidth]{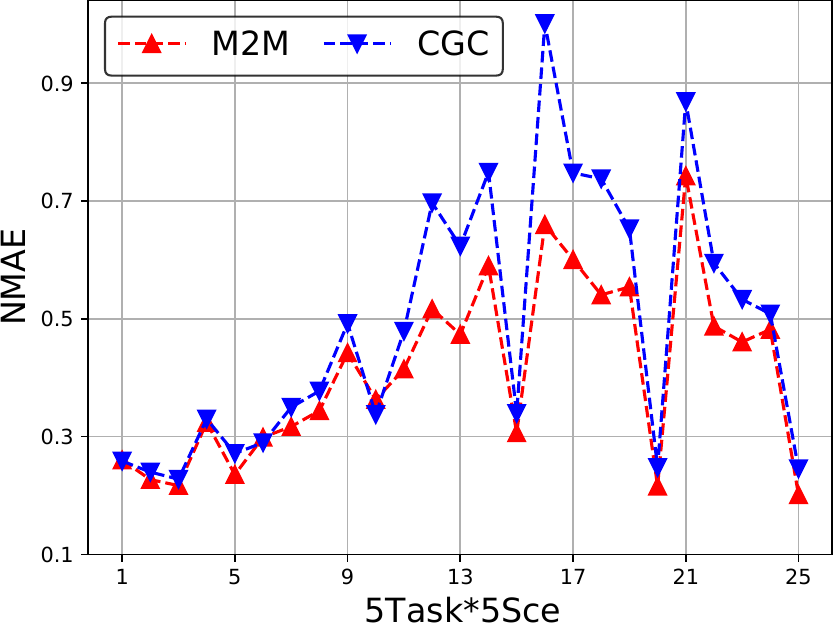}}
        \subfigure[{\footnotesize Overall Performance in SMAPE}]{\label{fig:jonitnpre_2}
        \includegraphics[width=0.48\textwidth]{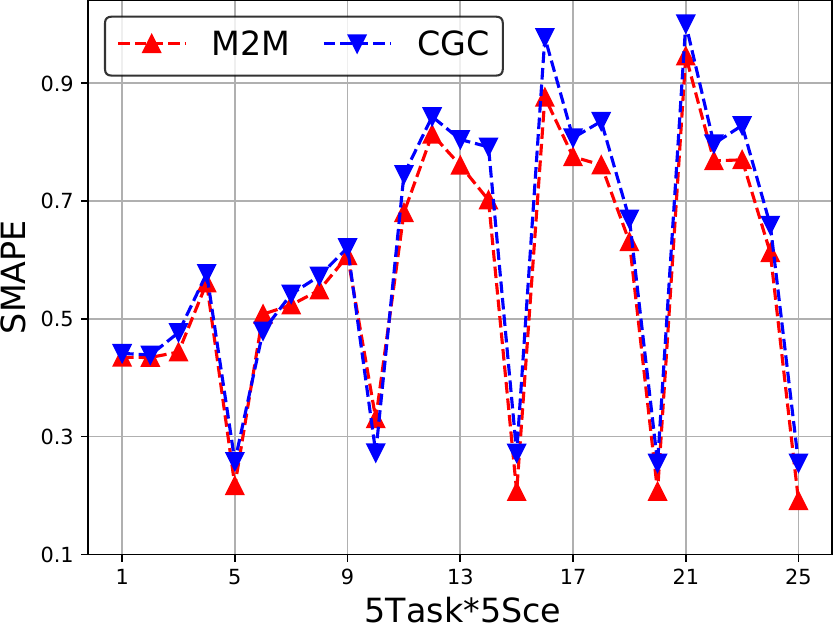}}
    \end{minipage}
    \vspace{-2ex}
    \caption{Overall Performance.} 
    \vspace{-5ex}
    \label{fig:overallperform}
\end{figure}
\subsection{Overall Performance Comparison}

We show experiment results in 
Table \ref{tab:Performance comparison}, where we find that our M2M outperforms other baselines consistently on different tasks from different scenarios. This demonstrates the effectiveness of our proposed method. Specifically, For Task Click respectively in scenarios A, B and C, our model achieves (+4.7\%, +24.1\%, +26.7\%) and (+6.9\%, +5.4\%,+8.9\%) improvements over the best baseline on NMAE and SMAPE. Similarly, the improvement of NMAE and SMAPE on Task Active-rate are (+13.3\%, +9.8\%, +12.8\%) and (+15.9\%, +24.1\%, +18.9\%) respectively. Moreover, we find that MMoE, CGC, Cross-Stitich methods outperform the single task learning approach, which demonstrates the effectiveness of learning multiple tasks jointly by exploiting relationships between tasks. Interestingly, we notice that improvements to M2M on Task Click turn out to be quite different from scenario to scenario. This might be related to the fact that scenario A has a larger scale, while smaller-scale scenarios B and C record larger improvements as M2M has the ability to transfer scenario knowledge via meta unit. 

Finally, the detailed performance is displayed in Figure \ref{fig:jonitnpre_1} and Figure \ref{fig:jonitnpre_2}, where we use NMAE and SMAPE to evaluate the overall performance of M2M with the best baseline CGC. Our M2M significantly outperforms CGC consistently on multi-scenario and multi-task.

\begin{table}[]
\caption{Results on the ablation study of the proposed M2M.}
\label{tab:ablation}
\small
\setlength{\tabcolsep}{0.5mm}{
\begin{tabular}{@{}c|cc|cc|cc@{}}
\toprule
\multirow{2}{*}{Model} & \multicolumn{2}{c|}{Task.click(Sce.a)} & \multicolumn{2}{c|}{Task.click(Sce.b)} & \multicolumn{2}{c}{Task.click(Sce.c)} \\
                      & NMAE               & SMAPE             & NMAE               & SMAPE             & NMAE               & SMAPE             \\ \midrule
\textbf{Full Model}    & \textbf{0.2168}    & \textbf{0.4435}   & \textbf{0.4733}    & \textbf{0.7605}   & \textbf{0.5406}    & \textbf{0.7611}   \\
\textbf{w/o MT}       & 0.2241             & 0.4612            & 0.5710             & 0.7923            & 0.6593             & 0.8158            \\
\textbf{w/o MA}       & 0.2189             & 0.4461            & 0.4966             & 0.7883            & 0.5768             & 0.7836            \\
\textbf{w/o TL}        & 0.2200             & 0.4493            & 0.5282             & 0.7612            & 0.5465             & 0.7643            \\\bottomrule
\end{tabular}}
\end{table}

\begin{figure*}
    \begin{minipage}{1.0\linewidth}
        \centering   
        \vspace{-10pt}
        \subfigure[{\tiny Temporal Length T}]{\label{fig:Temporal Sequence Length T nmae}
        \includegraphics[width=0.15\textwidth]{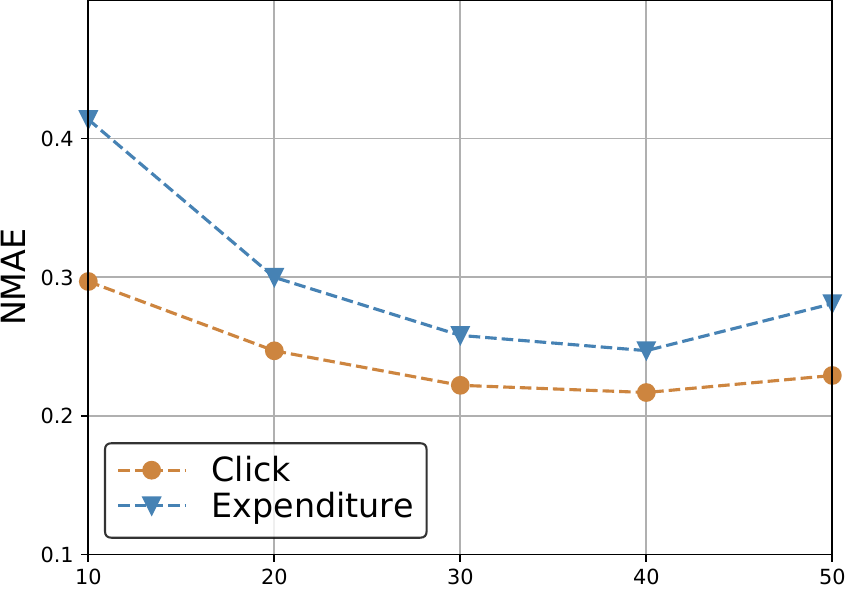}}
        \subfigure[{\tiny Temporal Length T}]{\label{fig:Temporal Sequence Length T smape}
        \includegraphics[width=0.15\textwidth]{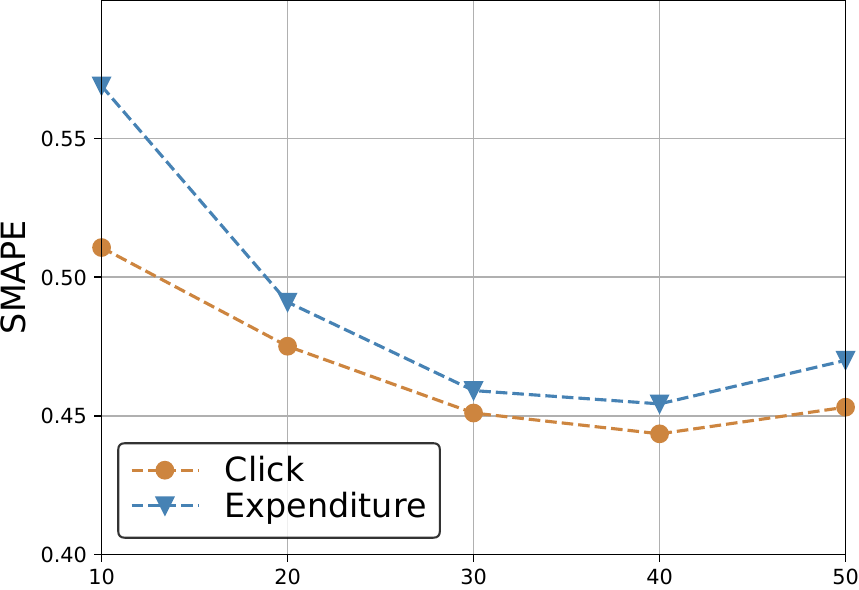}}
        \subfigure[{\tiny Scenario Knowledge Dimension d}]{\label{fig:Meta Dim d NMAE}
        \includegraphics[width=0.15\textwidth]{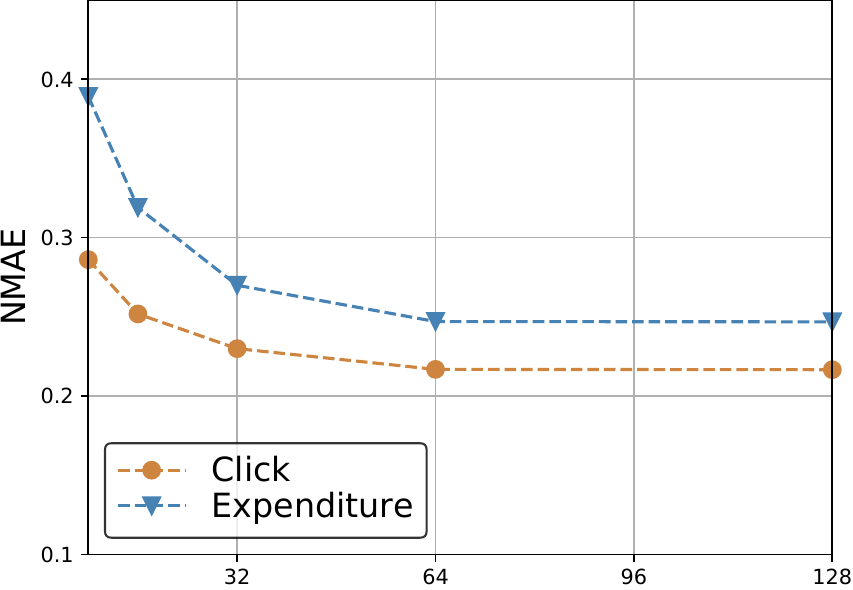}}
        \subfigure[{\tiny Scenario Knowledge Dimension d}]{\label{fig:Meta Dim d SMAPE}
        \includegraphics[width=0.15\textwidth]{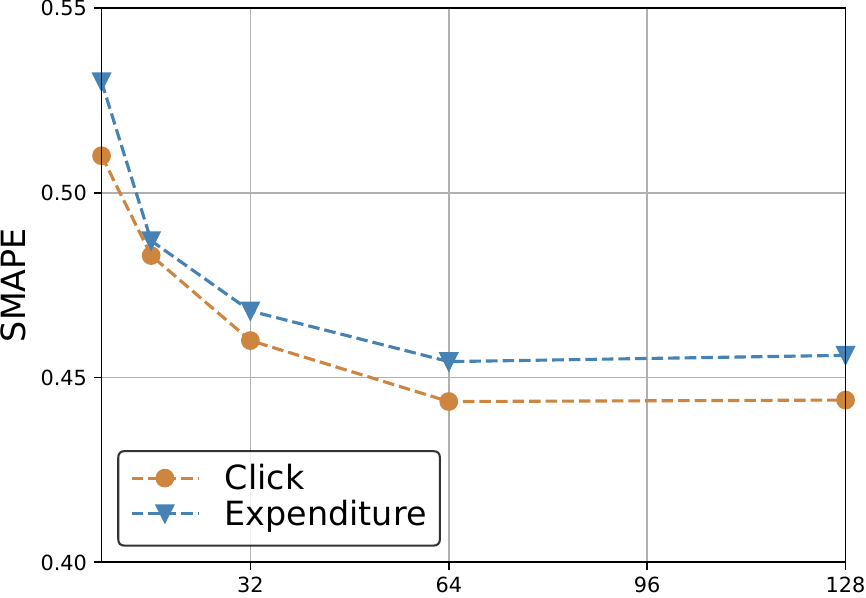}}
        \subfigure[{\tiny Meta Unit Depth K}]{\label{fig:Meta Unit Depth K NMAE}
        \includegraphics[width=0.15\textwidth]{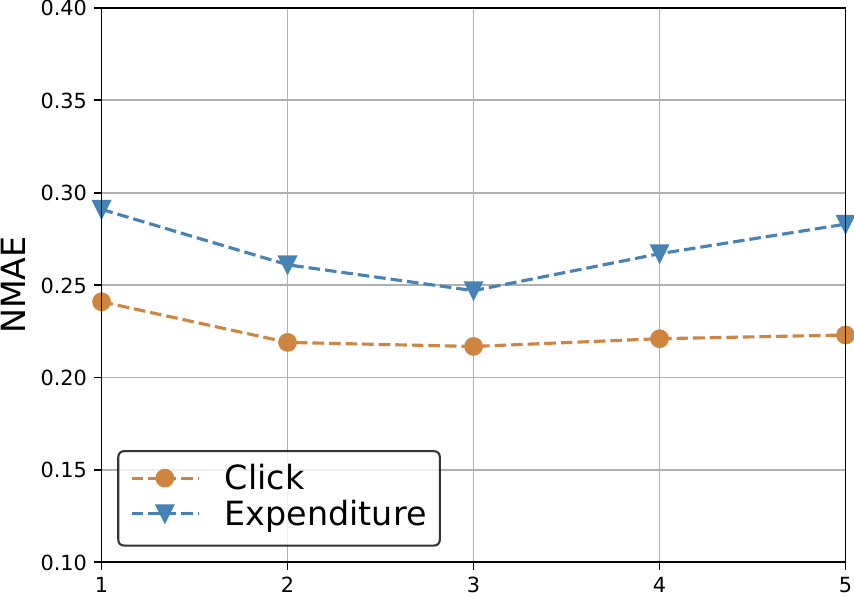}}
        \subfigure[{\tiny Meta Unit Depth K}]{\label{fig:Meta Unit Depth K SMAPE}
        \includegraphics[width=0.15\textwidth]{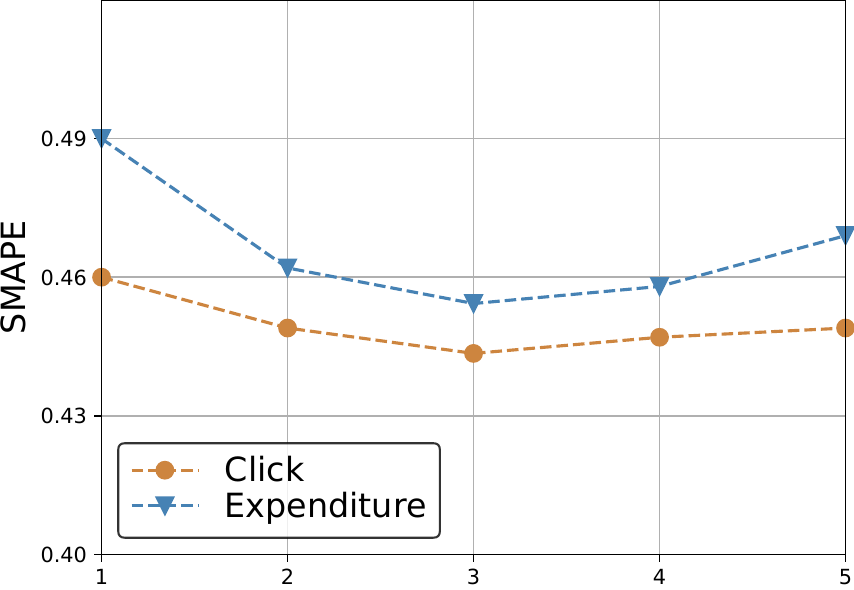}}
    \end{minipage}
    \vspace{-3ex}
    \caption{Hyperparameter Sensitivity.} 
    \vspace{-3ex}
    \label{fig:sensi}
\end{figure*}

\subsection{In-depth Analysis}
To better understand the performance of the proposed model, we conduct a series of in-depth analysis on M2M.
\subsubsection{Ablation Analysis}
To validate the contribution of each component of M2M, we conduct an ablation study by examining the performance after removing each component, listed and denoted as follows.
\begin{itemize}
    \item \textbf{M2M(Full Model)}: Use all components. 
    \item \textbf{w/o meta tower module (MT)}. Simple version without learning the high-order interactive features for a specific scenario.
    \item \textbf{w/o meta attention module (MA)}. Simple version without capturing diverse inter-scenario correlations.
    \item \textbf{w/o transformer Layer (TL)}. Simple version without transformer layer.
\end{itemize}

Table \ref{tab:ablation} shows the effectiveness of each module. And results are summarized as following: (1) The meta learning mechanism which contains meta attention module and meta tower module plays an important role in M2M. As shown, M2M without meta tower module and M2M without meta attention module are inferior to M2M. This proves that the meta attention module and meta tower module are effective in identifying lower-level fusion patterns and higher-level mixture patterns for specific scenarios respectively. (2)
As shown, M2M without transformer layer is inferior to M2M. This proves that transformer layer can effectively model the the time-series dependency. 

\subsubsection{Hyperparameter sensitivity}
In the following studies, we learn the impact of key hyper-parameters including the temporal sequence length T, the dimension of scenario knowledge d, and the number of neural network layers K in our meta unit one at a time by holding the remaining hyper-parameters at the optimal settings. The evaluation results in predicting click and expenditure tasks with SMAPE and NMAE metrics are shown in Figure \ref{fig:sensi}. The major findings are summarized as follows :

Figure \ref{fig:Temporal Sequence Length T nmae} and figure \ref{fig:Temporal Sequence Length T smape} presents the results of NMAE and SMAPE with different temporal sequence length T, respectively. The metrics reach optimal when the sequence length is 40 and the performance decreases when the sequence becomes longer. This can be explained by the reason that a longer sequence length tends to introduce extra information and more noises. The dimension of scenario knowledge d is another important parameter. To check the sensitivity of d, we plot overall NMAE and SMAPE of M2M with d = 8, 16, 32, 64 and 128 in Figure \ref{fig:Meta Dim d NMAE} and \ref{fig:Meta Dim d SMAPE}, with the increase of d, the model gets better performance. After d increases to 64, the result of model prediction is relatively stable. Figure \ref{fig:Meta Unit Depth K NMAE} and \ref{fig:Meta Unit Depth K SMAPE} show the NMAE and SMAPE for the different numbers of layers K in the meta unit. From this figure, we find that deeper networks can not guarantee to achieve better results. The model achieves its best performance when K is 3. When the network becomes deeper, the performance deteriorates. This is due to the reason that a deeper network is more likely to suffer from overfitting issues.

\graphicspath{{fig/}}
 \begin{figure}
   \centering 
   \vspace{15pt}
   \includegraphics[scale=0.12]{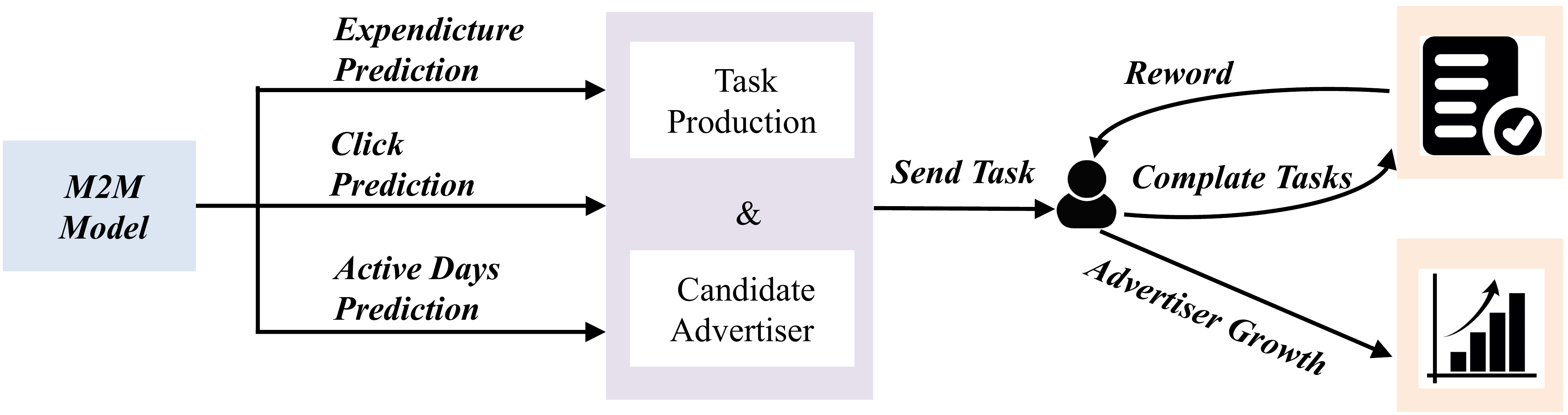}  
   \vspace{0pt}
   \caption{ 
    The pipeline of advertiser modelling and rewarded task tool application. }\label{fig:taskpip}
   \vspace{-3mm}
 \end{figure}

\subsection{Online A/B Test}

The proposed M2M method for advertisers modeling builds the foundation for many downstream applications such as Rewarded Task and Push Message which aim to improve the engagement and retention of advertisers in Taobao ecosystem. As Figure \ref{fig:taskpip} demonstrates, with our M2M model, we obtain the predictions of expenditure, clicks and active days in next 7 days, and then select the candidate advertisers whose future active rate is low; Next, the Task Production module assigns personalized tasks on top of those predictions. Selected advertisers are motivated to accomplish these tasks in order to get rewards.



Table \ref{tab:onlinetest} presents the results of online AB tests on the Rewarded Task Tool. Between 2021-March-11 and 2021-March-29, we conducted several rounds of online A/B tests in different scenarios. Take sponsored search scenario for example, the new M2M model and Single Task base model generated equal size of predicted future-inactive advertisers respectively as test and base group, about 100,000 each. With the same type of incentive tasks and incentive budget applied, we observed an improvement of +2.59\% in terms of the active rate, +2.09\% in terms of ARPU(Average Revenue Per User). This online experiment demonstrates the effectiveness and practicability of our model in the industrial landscape.

\section{Conclusion}

\begin{table}[]
\vspace{10pt}
\caption{A/B Test of the proposed model compared to the base model.}
\label{tab:onlinetest}
\setlength{\tabcolsep}{8.1mm}{
\begin{tabular}{c|c}
\hline
Indicators       & Gain             \\ \hline
Active Rate      & +2.59\%           \\
Average Revenue Per User             & +2.09\%           \\\hline
\end{tabular}}
\vspace{2pt}
\end{table}

In this paper, we proposed an efficient and scalable framework named M2M for multiple-scenario multiple-task advertiser modeling. First, a backbone network is adopted to learn representations of advertiser-related features and tasks. Then, to learn the meta-knowledge of scenario-related information explicitly, we design a Multi-level Meta Unit with customized scenario-related input, and its output is used to generate layer weights for later prediction networks. Furthermore, to combine multiple-scenario multiple-task in an end-to-end fashion effectively, we propose a Meta Learning Mechanism, which comprises a meta attention module and a meta tower module, respectively to capture diverse scenario-task correlations and enhance scenario-specific feature representation. Finally, we utilize the joint Poisson loss for multiple tasks to optimize our model in an end-to-end manner. Compelling results from both offline evaluation and online A/B tests demonstrate the superiority of M2M over state-of-the-art methods, especially in new and minor marketing scenarios. For future work, we'd like to incorporate more types of tasks like classifier beyond regression on top of the prediction network. Besides, it might also be interesting to combine our M2M model with optimization-based meta learning approaches like MAML, Reptile etc. 

\newpage


\bibliographystyle{ACM-Reference-Format}
\balance
\bibliography{ref}

\end{document}